\documentclass[journal]{IEEEtran}

\ifCLASSINFOpdf
\else
\fi
\usepackage{balance}

\usepackage{graphicx}
\usepackage{booktabs}
\usepackage{cite}
\usepackage[dvipsnames,svgnames,x11names,hyperref,table]{xcolor}
\usepackage{amsfonts}

\usepackage{amsmath}
\usepackage{multirow}
\usepackage[linesnumbered,ruled,lined,boxed]{algorithm2e}
\DontPrintSemicolon 

\SetCommentSty{mycommfont}
\usepackage{setspace}
\usepackage{subcaption}

\newcommand{\beginResultsTable}[7]{
}

\begin{document}
 \title{Fast-FNet: Accelerating Transformer Encoder Models via Efficient Fourier Layers}


\author{Nurullah Sevim,
        Ege Ozan Ozyedek,
        Furkan \c{S}ahinu\c{c},
        and~Aykut Ko\c{c},~\IEEEmembership{Senior Member,~IEEE}
\thanks{N. Sevim and A. Ko\c{c} are with the Department of Electrical and Electronics Engineering, Bilkent University, 06800 Ankara, Turkey, and also with the UMRAM, Bilkent University, Ankara, Turkey. E. O. Ozyedek is with the KU Leuven, Leuven, Belgium. F. \c{S}ahinu\c{c} is with the TU Darmstadt, Darmstadt, Germany. e-mail: aykut.koc@bilkent.edu.tr}
}

\markboth{}%
{Shell \MakeLowercase{\textit{et al.}}: Bare Demo of IEEEtran.cls for IEEE Journals}

\maketitle

\begin{abstract}
Utilization of the attention mechanism in Transformer-based language models leads to substantial performance improvements in almost all natural language processing (NLP) tasks. Similar attention structures are also extensively studied in computer vision models. Although attention mechanism enhance the model performance significantly, since it has quadratic complexity, the attention mechanism becomes inefficient for processing long sequences. Some recent studies show that Transformer models can still reach competitive results without the attention layer. One of this studies (FNet) suggests to replace attention layer with Fourier Transform (FT) in the Transformer encoder architecture and achieves competitive results with the original Transformer encoder model while accelerating training by removing the computational burden of attention. However, the FNet model ignores important FT properties that can be leveraged to further increase model efficiency. We propose different methods to deploy FT efficiently in Transformer encoder models. Our proposed architectures have shorter training times and have additional performance improvements in downstream tasks.
\end{abstract}


\IEEEpeerreviewmaketitle

\section{Introduction}\label{sec:intro}


\IEEEPARstart{T}{he} introduction of transformers \cite{vaswani2017attention} with specialized attention layer drastically improved the performance of neural network architectures for many machine learning tasks, mainly including language and image related frameworks \cite{devlin2019bert,zhou2021topicbert,huang2021aec-lstm,cui21pretraining,kaselimi22vision,an22domain}. Offering a great deal of performance enhancements, Transformers became quite popular and many deep learning benchmarks were challenged using Transformer-based architectures. Although Transformer-based models bring downstream task performance to upper levels, they are still open for improvements in terms of task performance and efficiency \cite{devlin2019bert,radford2019gpt2,brown2020gpt3,bai21fast}. Specifically, the model size and enormous training costs of Transformer models lead to the introduction of several other approaches, aiming to bring solutions to these computational efficiency problems. For instance, parameter sharing \cite{lan2020albert}, distillation, \cite{sanh2019distilbert} or modifications in the training objective \cite{liu2019roberta,wang22aroberta} are popular and effective improvements. In the meantime, some studies investigate the effects of deprivation of specific components of the Transformer such as the attention layer and positional encodings \cite{zhang19attention,ma19global,you2020hard,zheng20improving,tay2021synthesizer,haviv2022positional}. 

While improvement in model efficiency is an important technical objective itself, it is also increasingly important from other perspectives such as decreasing the carbon footprint of deep learning models. Recent studies show that using deep models that require expensive computational resources may have adverse environmental effects \cite{henderson20towards,bender21danger,patterson22carbon}. \cite{patterson22carbon} suggests that through using more efficient deep learning models, such as sparse models instead of dense models, the computation costs drop and eventually the carbon footprints of machine learning algorithms decline. Therefore, our efforts to decrease the computational cost of Transformer models also lead to positive environmental effects.

One way to increase the efficiency and even enhance the information capturing capability of deep learning models is to use signal processing tools. Specifically, Fourier transform (FT) and wavelets are such tools that can provide performance and efficiency improvements when using them in neural networks architectures \cite{mathieu14fast,vasilache2015fbfft,zhang2013fault,tamkin2020prism,lee2021fnet}. Recently the fractional versions of these aforementioned transforms have also started to involve in the neural network architectures either as feature extractors or as learning blocks \cite{liu19fractional,shi2021wavelet, zhao22fracimage}. In the Prism model that is introduced by \cite{tamkin2020prism}, spectral filters are applied to the activations of neurons to capture the features of different scaled parts of language, such as word-level, utterance-level, and document-level. These spectral filters are analogous to conventional low-pass, high-pass, and band-pass filters. In their pioneering work, \cite{lee2021fnet} brings another perspective to the Transformer encoders by replacing attention layers with FT to propose the FNet model. With this replacement, FNet can be trained $80\%$ faster than the original BERT model by maintaining $92\%$ of the accuracy performance in GLUE tasks \cite{wang2018glue}.

Although FNet can preserve model performance to some extent in the absence of attention layer, it does not exploit the full potential and properties of the FT, which are essential in classical signal processing \cite{vetterli14signal}. For example, FNet takes only real parts of the 2D Fourier Transform of the given input embeddings. This choice results in two basic complications. First, since the FT of a real signal is conjugate symmetric, the real part of the output from the FT has a symmetric structure. Therefore, redundancies occur when feeding the output of FT to the next layer, which degrades both the performance and efficiency. Second, using only the real part of the Fourier transformed input, the full characteristics of the data cannot be conveyed throughout the model architecture.

In this study, we present a full-scale work that incorporates FT to the Transformer encoder to propose Fast-FNet. We propose multiple solutions consisting of several pooling methods and adding multilayer perceptron (MLP) layers to remedy aforementioned disadventages of the FNet. We implement exhaustive experiments to observe the effects of the proposed methods. We gain considerable computational efficiency compared to the FNet without losing accuracy. Our main contributions can be summarized as follows:
\begin{itemize}
    \item Proposed Fast-FNet with a more efficient Fourier layer utilizing properties of FT.
    \item Proposed modifications to Transformer model architecture to match the input size with newly proposed Fourier layer.
    \item Proposed efficient Transformer model architectures in terms of both training speed and memory requirements.
\end{itemize}

The rest of the manuscript is organized as follows. In Section \ref{sec:related}, related work is presented. Our proposed Fast-FNet architecture is introduced in Section \ref{sec:methodology}. In Section \ref{sec:exps}, experimental details are given and obtained results are discussed. Finally, we conclude our work in Section \ref{sec:conclusion}.

\section{Related Work}\label{sec:related}

\subsection{Attention Mechanism}
Many deep language models show substantial improvements on downstream task performances by utilizing attention mechanism \cite{devlin2019bert,radford2019gpt2,brown2020gpt3, belainine2022dialogue, zhou2021topicbert, liu2021empirical}. Capturing contextual relations of given data sequence makes attention mechanism to be appealing for not only fundamental natural language processing (NLP) tasks but also other deep learning tasks. For example, \cite{ni2021comparative} uses multi-attention-based convolutional neural networks (CNNs) to extract features. Multi-attention includes self-attention and cross attention. Self-attention considers the attention within users and items while cross-attention considers the attention between users and items \cite{ni2021comparative}. Attention is also used in tasks dealing with time series data \cite{agrawal2022score}. In computer vision applications, vision Transformer model is also proposed \cite{dosovitskiy2021vit}.

Attention mechanism can be adapted various learning schemes apart from the Transformer architecture. For instance, \cite{zhang2020gru} proposes gated attention model as a novel gated recurrent unit (GRU) for neural machine translation. \cite{huang2021aec-lstm} also integrates attention mechanism with long short-term memory (LSTM)  networks for sentiment analysis. In \cite{ni2021comparative}, CNNs are used along with multi-attention mechanism in a recommendation model. 

Several studies are conducted to obtain further understanding of Transformer-based models along with attention mechanism and its effectiveness in different layers \cite{tenney2019rediscover,ramsauer2021hopfield,lu22attention}. \cite{tay2021synthesizer} shows that replacing the matrix obtained by query and key operations in dot-product attention with randomly initialized matrix still yields competitive performance with regular self attention. In a similar study, all but one attention head of each encoder layer are replaced with non-learnable attentive patterns \cite{raganato2020fixed}, which are based on positional information and do not include and need any external knowledge. Neural machine translation experiments also show that fixing attention heads does not affect translation performance \cite{raganato2020fixed}. On the other hand, \cite{you2020hard} replaces all self-attention heads in encoder and decoder with fixed Gaussian distributions centered around a different token for each head. \cite{you2020hard} observes that this learning scheme has minimal effects on machine translation performance, and that performance decreases more severely when the replacement is implemented in cross-attention between encoder and decoder instead of self-attention \cite{you2020hard}. Since utilization of attention is not limited to NLP, similar studies trying to find alternatives to attention are also proposed in computer vision field. For example, \cite{tolstikhin2021mlp} proposes multi-layer-perceptron layers instead of using attention in Vision Transformer \cite{dosovitskiy2021vit} model or using conventional CNN architecture. MLP-mixer leads to no performance degradation in image classification tasks. In another study, fast multi-pole method (FFM) is utilized for linearization of the complexity of the attention calculation \cite{nguyen2021fmmformer}. GFNet model that can be considered as computer vision counterpart of the FNet and Prism model is proposed to solve image classification efficiently \cite{rao2021gfnet}. In GFNet, self-attention is replaced with three operations, Fourier transforming, applying learnable filters, and returning back to the original domain via inverse FT.

\subsection{Fourier Transform in Deep Learning}

FT is very-well known and constitutes one of the most essential tools of signal processing. The history of Fourier analysis in deep learning literature is also quite rich. Previously, several various versions of neural networks resembling Fourier Series or FT are introduced \cite{gallant1988there,silvescu1999fourier,liu2013fourier,gashler2014deep}. The general motivation of these efforts lies behind changing activation of the neurons so that models perform better and converge fast. There are also many instances that FT is utilized to enhance the performance of recurrent neural networks (RNNs) \cite{koplon1997fourier,zhang2000forenet,zhang2018fru,wolter2020spectral}. Main motivation of these studies is to capture sequential features or long-term dependencies that exist in data. In \cite{koplon1997fourier}, Fourier-type activation function is proposed to be used in RNNs for fitting sequential input/output data. Zhang \textit{et al.} apply Fourier analysis to build a complex valued RNN model such that time series analysis and recurrent networks can be integrated \cite{zhang2000forenet}. In \cite{zhang2018fru}, Fourier Recurrent Unit (FRU) is proposed as a solution for vanishing gradient problem. Wolter \textit{et al.} uses short-time Fourier transform (STFT) to efficiently process sequential data with RNNs \cite{wolter2020spectral}. They also apply parameter reduction by low-pass filtering.

Due to the close relation between convolution operation and FT that the convolution operation is simply equivalent to multiplication in Fourier domain, CNNs are very suitable structures to utilize Fourier features, especially in computer vision. Based on this concept, Mathieu \textit{et al.} \cite{mathieu2014fast} propose to multiply the feature maps of input images by corresponding kernels in Fourier domain instead of implementing convolution. There are also previous attempts to employ Fourier features in face detection task \cite{ben1999fast}. However, these attempts do not utilize FT in training phase. On the other hand, in \cite{pratt2017fcnn}, Fourier CNNs are proposed where the entire training procedure is implemented in Fourier domain. Main advantage of this model comes from efficient implementation of training without sacrificing task performance. Inspired by Fourier layer of the FNet model \cite{lee2021fnet}, Pan et al. propose Walsh-Hadamard transform block to replace the convolution layers \cite{pan2022hadamard}. Similarly, FNet architecture is utilized in energy disaggregation  \cite{nalmpantis2022energy}. However, authors take complex information in Fourier domain into consideration as a difference from FNet. In \cite{li2020falcon}, fast Fourier transform (FFT) properties are used in designing encryption algorithms to provide safe CNN predictions. There are also several other studies utilizing FT for efficient computation and performance increases \cite{elbakry2004fast,rippel2015spectral,khan2019dropout,highlander2015efficient,shi2021wavelet,vasilache2015fbfft,ryu2018dft,tancik2020high}.

In the literature, FT has been deployed not only as a part of network architecture but also as a feature extraction tool. Especially in signal processing field, detection and classification problem of electrocardiogram signals are approached by a combination of FT and neural networks \cite{minami1999real,gothwal2011cardiac,mironovova2015electrocardiogram}. \cite{tan2005aircraft,zhang2013fault} also exploit Fourier features in fault diagnostics. Control theory is also one of the fields integrating Fourier and neural networks. Iterative learning control scheme using the Fourier neural network (FNN) is presented to improve the tracking performance of control systems \cite{zuo2008control}. On the other hand, \cite{li2021fno} proposes the Fourier Neural Operators (FNO) to solve differential equations. Based on this study, models that improve the efficiency of existing Transformer models are also proposed \cite{guibas2022efficient,cao2021choose}.

Recently, the fractional form of FT, i.e., the Fractional Fourier Transform (FRFT), is also started to be integrated into neural network architectures. FRFT is a generalized version of FT with a fraction parameter, where the input signal can be mapped to a continuum of intermediate domains between time and frequency \cite{ozaktas01frt,ozaktas2001fractional}. In a pioneering work, \cite{zhao22fracimage} embeds FRFT into image Transformer architecture to extract global and contextual features from hyperspectral images (HSI) and light detection and ranging (LiDAR) data. Through this newly proposed architecture and the advantage of FRFT as a feature extractor the authors reported improvements in both performance and efficiency on benchmark tasks.

\subsection{Transformers in Signal Processing}

The Transformer architecture takes place in several signal processing related areas. For example, Liang et al. adapt the Transformer to reconstruct high-resolution light field images from low-resolution instances \cite{liang2022light}. Furthermore, studies focusing on skeleton-based action recognition start to exploit Transformer-based models in collaboration with attention mechanism \cite{gao2021skeleton,kong2022mtt}. In speech recognition task, Chen et al. propose non-autoregressive Transformer model with Audio-Conditional Masked Language Model to alleviate inference computation burden of auto-regressive counterparts \cite{chen2021nar}. Later in \cite{tian2022hybrid}, an hybrid model consisting of autoregressive and non-autoregressive Transformers are proposed for speech recognition. On the other hand, the Transformer is also integrated into communication systems. In \cite{xie2021communication}, an end-to-end deep learning-enabled semantic communication model is proposed. The model mainly consists of semantic encoder, channel encoder, channel decoder, and semantic decoder. Similarly, notion of Transformer is utilized in multiple-input multiple-output detection tasks in communications \cite{pratik2021mimo}. In \cite{scribano2022dct}, discrete cosine transform (DCT) is used to reduce the complexity of the dot-product attention calculations. Authors also show that data compression feature of DCT can be applied to query, key, and value matrices of attention layer of input embeddings. This method reduces training time while obtaining competitive performances compared to vanilla Transformer. With similar motivation, Nystr\"{o}m matrix approximation is used to approximate standard self-attention \cite{xiong2021nystromformer}.

\section{Fast-FNet}\label{sec:methodology}

\subsection{Preliminaries on Attention Mechanism}

Attention mechanism can be described as the mapping of a query (Q) and a collection of key (K) and value (V) pairs to an output \cite{vaswani2017attention}. This mapping is mathematically described as:
\begin{equation}
    \mathbf{Attention}(\mathbf{Q},\mathbf{K},\mathbf{V}) = softmax(\dfrac{\mathbf{Q}\mathbf{K}^T}{\sqrt{d_k}})\mathbf{V},
\end{equation}
where query, key, and value matrices $\mathbf{Q}$, $\mathbf{K}$, and $\mathbf{V}$ are learnable linear projections of input vector:
\begin{equation}
    \mathbf{Q} = \mathbf{XW_Q};\;  \mathbf{K} = \mathbf{XW_K};\;  \mathbf{V} = \mathbf{XW_V},
\end{equation}
where $\mathbf{W_Q}\in{\mathbb{R}^{d_{Model}\times d_k}}$, $\mathbf{W_K}\in{\mathbb{R}^{d_{Model}\times d_k}}$, and $\mathbf{W_V}\in{\mathbb{R}^{d_{Model}\times d_v}}$. Here $\mathbf{X}$ is the input data and $d_{Model}$ is the feature dimension of input.

The attention mechanism can be paralleled by projecting the queries, keys and values multiple times with different learnable matrices. Each projection of these matrices go through the attention operation to form `head's. Finally, all heads are concatenated and multiplied with a learnable weight matrix to generate the final output. This procedure is called Multi-head Attention (MHA) and can be described with the following equations:
\begin{equation}
    head_i = \mathbf{Attention(Q}_i,\mathbf{K}_i,\mathbf{V}_i\mathbf{)},
\end{equation}
\begin{equation}
    \mathbf{MHA(Q,K,V)}=Concat(head_1,...,head_h)\mathbf{W_O},
\end{equation}
where $\mathbf{W_O}\in{\mathbb{R}^{hd_v}\times d_{Model}}$.

\begin{figure}[h]
\begin{subfigure}{.3\textwidth}
  \centering
  \includegraphics[height = 4.4cm,width=5.8cm]{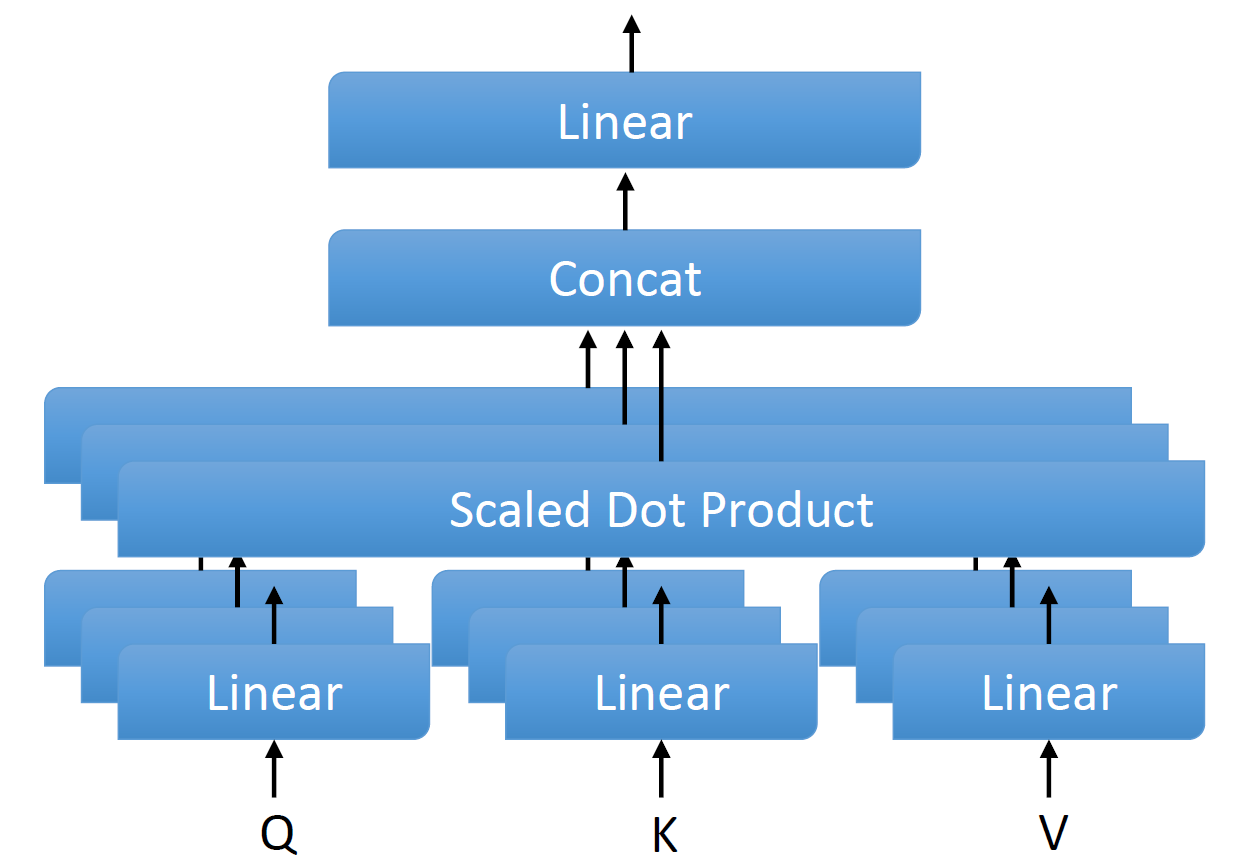}
  \caption{Multihead Attention Mechanism}
\end{subfigure}%
\begin{subfigure}{.2\textwidth}
  \centering
  \includegraphics[height = 4.4cm,width=3.6cm]{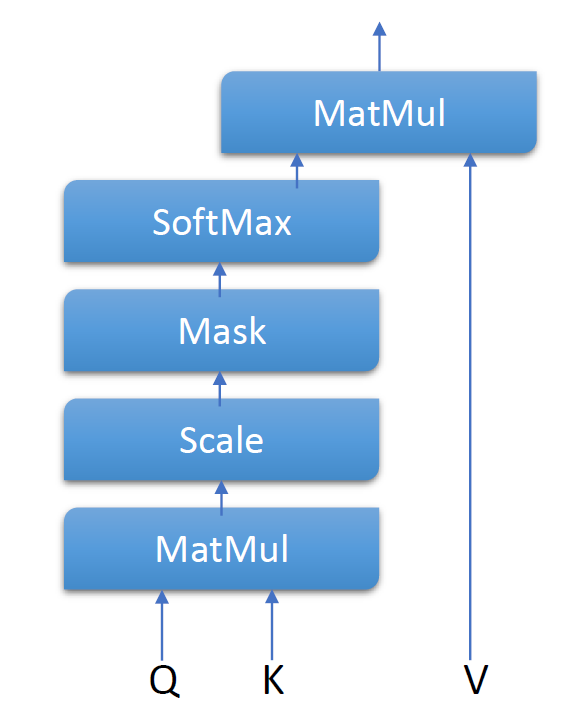}
  \caption{Scaled Dot Product}
\end{subfigure}
\caption{Multihead attention (MHA) mechanism and scaled dot product that is used in MHA \cite{vaswani2017attention}.}
\label{mha-sdp}
\end{figure}

One of the objectives for using the MHA is to use multiple GPUs for computations and fasten the model operations. A simple representation of this mechanism is given in Fig. \ref{mha-sdp}.

\subsection{FNet Model Details}

The FNet architecture is built in similar fashion to other NLP models, such as BERT \cite{devlin2019bert}, where the foundation of the model consists of the Transformer encoder blocks. In FNet, the attention layer is replaced with the discrete Fourier transform (DFT) to accelerate processing speeds while obtaining quality features that display accurate results \cite{lee2021fnet}. Only the real-valued output of the DFT operation is used to ensure that the feed-forward layer structure in each encoder block is unchanged. The output of the DFT layer can specifically be expressed in the form:
\begin{equation} \label{FNetEquation}
    y = \mathbb{R}(\mathbb{F}_{d_S, d_H}(x_{m,n})),
\end{equation}
where $\mathbb{F}_{d_S, d_H}$ is the 2D DFT operation along the sequence and hidden dimension for the given input. This serves a simple yet effective solution to the aforementioned problem of gaining speed without costing accuracy. Their results show that the FNet model is $40\%$ to $70\%$ faster than the BERT-Base model. Although the accuracy difference between the two models is apparent, FNet still proves to be quite successful in learning the GLUE task. This makes the FNet model a solution for applications where computational efficiency holds more importance than accuracy.

\subsection{DFT and its Conjugate Symmetry Property}

An essential part of our proposed Fast-FNet is the dimension reduction of the DFT output. For this reduction, we leverage the conjugate symmetry property of the FT. For a given $M\times N$ dimensional input $\{x_{m, n}\}^{M-1, N-1}_{m=0, n=0}$, the 2D-DFT is given as:
\begin{equation} \label{DFTformula}
    X_{u, v} =  \sum_{m = 0}^{M-1} \sum_{n = 0}^{N-1} x_{m, n} \: e^{-j2\pi (\frac{mu}{M} + \frac{nv}{N})} ,
\end{equation}

When the input sequence is real-valued, $x_{m, n} = \mathbb{R}(x_{m, n})$ for all $m$ and $n$, the DFT output is conjugate symmetric over both dimensions. The dimension reduction used for the Fast-FNet occurs over the hidden dimension, hence the output matrix now supports:
\begin{equation} 
    X_{u, v} =   X_{u, N-v}^*.
\end{equation}
Keeping in mind that the FNet architecture only uses the real parts of the DFT output as displayed in Eq.~\ref{FNetEquation}, conjugate symmetry property leads to the realization below:
\begin{equation} \label{EqualHalves}
    \mathbb{R}(X_{u, v}) =  \mathbb{R}(X_{u, N-v}^*),
\end{equation}
Hence, Eq. \ref{EqualHalves} essentially states that the output sequence the DFT layer in the FNet produces results in a sequence that has near-identical halves. It can only be defined as near-identical, since $X_{u, v}$ for $v = 0$ and $v = \frac{N}{2}$ are not equal to each other. As such, the same information is fed to the model twice, creating a redundancy. The proposed Fast-FNet model eliminates this property.

\subsection{Fast-FNet Architecture}
\begin{figure}[!ht]
    \centering
    \includegraphics[width=\columnwidth]{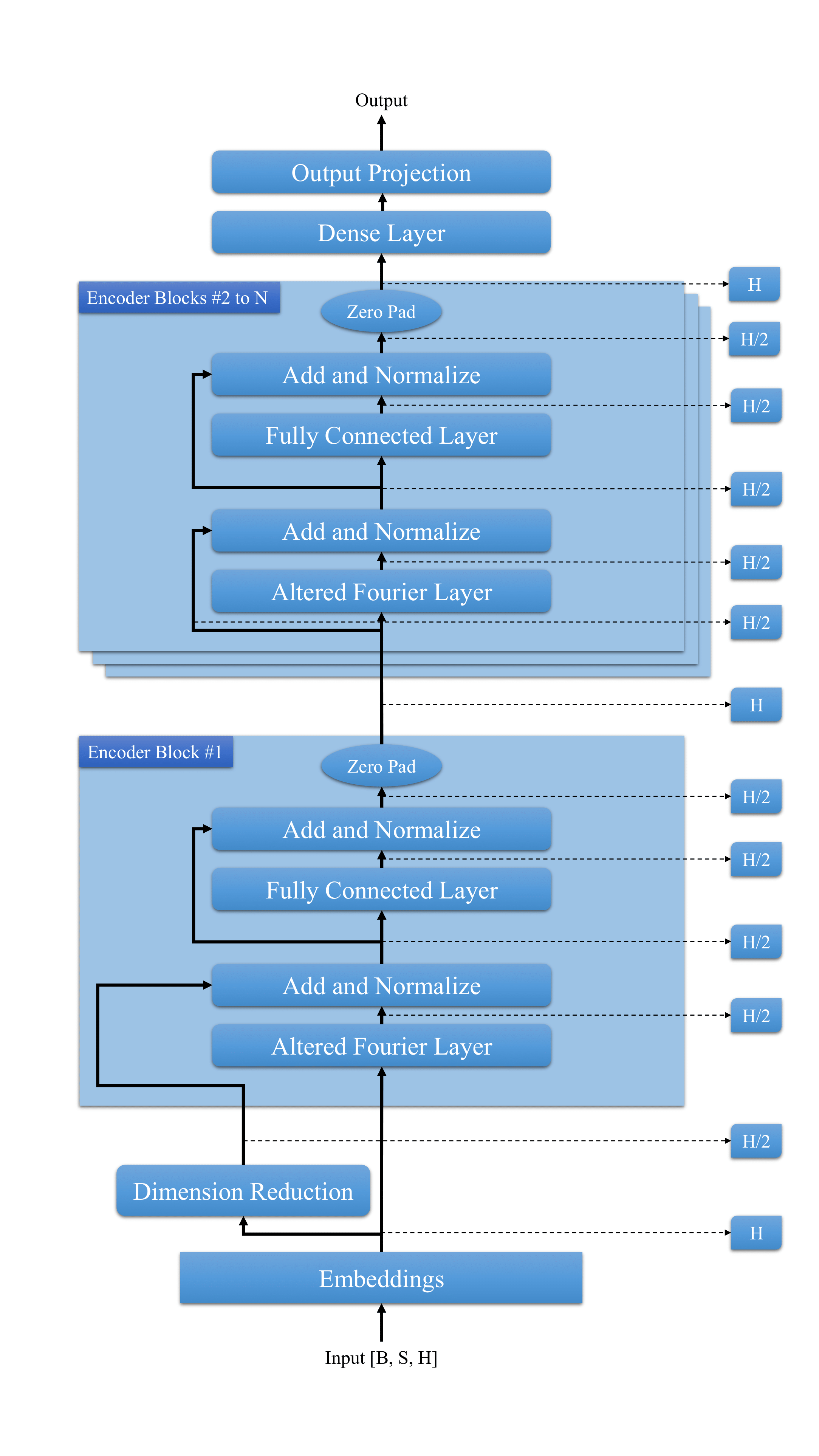}
    \caption{A schematic of the Fast-FNet architecture. Here, the input has dimensionality of $[Batch Size, Sequence Dimension, Hidden Dimension]$ i.e $[B,S,H]$. There are dimension indicators over the right side of the figure, with $H$ being the initial hidden dimension and $H/2$ being half of it. The first encoder block is fed both the reduced embeddings (for the residual) and the initial embeddings (for use in the Fourier Layer). The Fourier Layer is exchanged for an altered version, where the output is the first half of the DFT. Before the end of each encoder block, the output is zero padded to keep the hidden dimension stable. For all remaining encoder blocks, the added zeros are taken out to get the residual, while the zero padded version is used as the DFT input.}
    \label{FastFNetArchitecture}
\end{figure}

Following Eq. \ref{EqualHalves} and the explanation above, the DFT output sequence is cut in half and only the first half ($0 \leq k < N/2$) is used for the subsequent operations. The hidden dimension of the model reduces from $H$ to $H/2$. It should be noted that by doing this, we use all unique $X_k$ values except for $k = N/2$, which is a necessity to keep length of the sequence a power of $2$. 

Reducing the hidden dimension creates issues in the Transformer architecture that needs to be addressed. The first issue is that \textit{the Add and Normalize} layer cannot operate with two different input dimensions. To fix this, the input of the first encoder block has to go through dimensionality reduction. Then, the reduced version of the embeddings can be used in the add and normalize layer, while the untouched embeddings are connected to the DFT layer. Another issue arises for the subsequent encoder blocks. Since the output hidden dimension of the encoder block will be $H/2$, for continuity reasons the input size of the next encoder block should be the original hidden dimension, otherwise the network will shrink the dimension to half in every encoder block, which is not desired. To preserve the dimesionality of output tokens and not add any costly operations, the output of each encoder block is zero padded to the original hidden dimension $H$. The succeeding blocks use the entire zero-padded input for the DFT layer, and only the non-zero part is used in the Add and Normalize layer. The entire process that takes place in a single encoder block is mathematically described as follows:
\begin{equation}
    \mathbf{X_{AFT}} = \mathbb{R}(\mathbb{F}_{dS,dH}(\mathbf{X}))_k,\;k=0,...,h/2.
\end{equation}
\begin{equation}\label{dimred}
    \mathbf{X_{norm1}} = Norm(\mathbf{X_{AFT}}+DimRed(\mathbf{X})),
\end{equation}
\begin{equation}
    \mathbf{X_{FF}} = FF(\mathbf{X_{norm1}}),
\end{equation}
\begin{equation}
    \mathbf{X_{norm2}} = Norm(\mathbf{X_{FF}}+\mathbf{X_{norm1}}),
\end{equation}
\begin{equation}
    \mathbf{X_{OUT}} = ZeroPad(\mathbf{X_{norm2}}),
\end{equation}
where $\mathbb{F}_{d_S, d_H}$ is the 2D DFT operation along the sequence and hidden dimension, $Norm(\cdot)$ is the batch normalization, $DimRed(\cdot)$ is the dimension reduction method that is used, $FF(\cdot)$ stands for feed-forward layer, and $ZeroPad(\cdot)$ is the padding operation. The dimension reduction in Eq. \ref{dimred} only takes part before the first encoder block. For the rest of the encoder blocks, we only take the padded zeros out before the Add and Normalize stage. Note that the input of the FT layer except the one in the first encoder layer is still zero padded. Finally, the proposed Fast-FNet architecture can be visually seen in Fig. \ref{FastFNetArchitecture}.


\begin{algorithm}
\setstretch{1.2}
\caption{Fast-FNet Architecture}
\label{fastarc}
\KwIn{Word Embeddings x} 
\KwOut{Classification Output y} 
        $r \leftarrow \text{DimensionReduction}(x)$ \tcp*[r]{first residual} 
        \For{i $\leftarrow$ 0 \textbf{to} N-1}{
            $X \leftarrow  \mathbb{R}\{DFT(x)\}$ \;
        $X \leftarrow X \text{ from 0 to } H/2$ \tcp*[r]{identical halves} 
        $\hat{X} \leftarrow \dfrac{(X + r) - \mu_1}{\sigma_1}$ \tcp*[r]{normalize} 
        $I_1 \leftarrow W_{LN1} \hat{X} + b_{LN1}$ \tcp*[r]{1st layer} 
       $F \leftarrow \text{ReLU}(W_{F1} I_1 + b_{F1})W_{F2} + b_{F2}$ \;
       $\hat{F} \leftarrow \dfrac{(F + I_1) - \mu_2}{\sigma_2}$ \tcp*[r]{normalize} 
        $I_2 \leftarrow W_{LN2} \hat{F} + b_{LN2}$ \tcp*[r]{2nd layer} 
        $x \leftarrow  \begin{cases}
    I_2,& \text{if } 0 \leq d_h < H/2 \\
    0,              & \text{otherwise}
\end{cases} $\tcp*[r]{zero pad}     
            $r \leftarrow x \text{ from 0 to } H/2$\;
        
        }
    $I \leftarrow Wx+b$ \;
    $y \leftarrow \text{OutputProjection}(I)$     
        
\end{algorithm}

The algorithmic foundations of the general architecture of Fast-FNet and the operations that take place in a single encoder block of Fast-FNet are given in Alg. \ref{fastarc}. The variables $\mu$ and $\sigma$ that are used in the Add and Normalize layer are the mean and standard deviation of the input layer, respectively. Throughout the algorithm, dimension reductions only occur over this dimension, i.e., Line 4 of Alg. \ref{fastarc} (the token dimension is constant, but the hidden dimension is halved). Similarly, the embedding reduction and zero padding occurs over the hidden dimension as well. 

\section{Experiments \& Results}\label{sec:exps}

\subsection{Experimentation Details}

We tested our proposed Transformer models and the FNet with two benchmark frameworks. First, we used a common transfer learning framework for Transformers where we pretrain models with C4\footnote{https://github.com/google-research/text-to-text-transfer-Transformer\#c4} dataset and then fine-tune for GLUE \cite{wang2018glue} tasks. Experimentation is done following the BERT-Small and BERT-Medium specifications \cite{devlin2019bert}. The following naming convention is followed: [Model]-S refers $L = 4$ (encoder blocks) and $H = 512$ (hidden dimension), [Model]-M refers $L = 8$ (encoder blocks) and $H = 512$ (hidden dimension), and [Model]-Base refers $L = 12$ (encoder blocks) and $H = 768$ (hidden dimension), [Model] stands for the deep learning models that are used in the experiments, i.e FNet-S or Fast-FNet MaxP-M. The results of our proposed model Fast-FNet and the baseline FNet models have been obtained over training on 2 GeForce RTX 2080 Ti GPUs, both for training and classification. For all models, the batch size was $64$ and the feed-forward network dimension was set to be $4\times H$. Finally, the maximum sequence length was the same as the hidden dimension, which is $512$. 

The second benchmark is the Long Range Arena (LRA) \cite{tay2021lra}, which is introduced for comparing the efficiency of Transformer models. For LRA tasks, we used the code base in \cite{chen21skyformer}, where the benchmark tasks are implemented in PyTorch. The code base is obtained from the GitHub repository\footnote{https://github.com/pkuzengqi/Skyformer.git} and the configuration provided in the repository is used. For the training and testing phases of this experiment, we used a GeForce GTX 1080 Ti graphic card.

The experimentation for our models revolves around dimension reduction methods that are applied after the embedding layer. Not to diminish the efficiency gained by the altered DFT output while also obtaining a meaningful representation of the initial embeddings, three dimension reduction techniques, two pooling approaches and one dense network layer, are used.

\subsubsection{\textbf{Dimendsion Reduction Using Pooling}}

We used two pooling methods to reduce the embedding layer output to half of its dimension. Pooling methods use arithmetic operations and hence are uncostly both in terms of operation efficiency and memory space allocation. The first pooling method used is the max pooling, which can be expressed as:
\begin{equation} \label{maxpool}
    Y_k =  \begin{cases}
    X_{2k},& \text{if } X_{2k} > X_{2k + 1} \\
    X_{2k+1},              & \text{otherwise.}
\end{cases} 
,\;k=0,...,H/2.
\end{equation}
The second pooling method used is the mean pooling with the following formulation:
\begin{equation} 
    \label{meanpool}
    Y_k =  \dfrac{X_{2k} + X_{2k + 1} }{2},
\end{equation}

\subsubsection{\textbf{Dimension Reduction Using a Dense Network Layer}}
To reduce the hidden dimension to half, we also consider adding a learnable linear network layer to the architecture. 
\begin{equation} 
    \label{meanpool2}
    Y = W X + b,
\end{equation}
where $W$ is the learnable weight matrix and $b$ is the bias term. This is, in theory, a better choice for a reduction than pooling methods since it is learnable. However, it adds computational cost.




\subsection{Results}
\subsubsection{\textbf{Transfer Learning and GLUE Tasks}}
The parameter sizes are reported following \cite{devlin2019bert} with the slight difference that is the vocabulary model as noted in \cite{lee2021fnet} (we followed the vocabulary size of the FNet, which is $32,000$ compared to BERT's $30,522$). The reduction caused by the adjustment in Fourier layers only affects the parameter size of the encoder blocks. Since, in a single encoder layer, DFT reduces the dimension to its half, the parameter sizes of the following feed-forward layer and layer normalization stages are also reduced to the half ($49.6\%$ reduction when biases are considered).

In Table \ref{paramsize}, the parameter sizes for different model specifications are given. For smaller models, the FNet-S and Fast-FNet-S have similar parameter sizes although a $16\%$ reduction is present. The difference between parameter sizes of the FNet and our proposed models becomes clearer when the number of encoder blocks increases. For the [Model]-M models, for example, the reduction of parameter size for our proposed models goes around $24\%$ and for even larger models, such as the [Model]-Base specification, the difference in parameter size is notable and shows around $34\%$ reduction.

\begin{table}[htb]
\centering 
\resizebox{\columnwidth}{!} {%
\begin{tabular}{l@{\extracolsep{4pt}}ccc}
\toprule   

\multirow{3}{*}{\textbf{Model}}  & \multicolumn{3}{c}{\textbf{Parameter Size} (Millions)}\\

 & -S & -M & -Base\\
  & L = 4, H = 512 & L = 8, H = 512 & L = 12, H = 768\\
\midrule
FNet &  25.3 & 34.0 & 82.3\\ 
Fast-FNet with Pooling & 21.1 & 25.6 &  53.9\\ 
Fast FNet with added Dense Layer & 21.3 & 26.0 & 54.2\\
\bottomrule
\end{tabular}
}
\vspace{0.5em}
\caption{Parameter sizes for the FNet and the proposed models. The number of encoder layers are indicated by $L$ and the hidden dimension is indicated by $H$.} 
\label{paramsize}
\end{table}

Since the matrix multiplications become faster, the reduction in parameter sizes also causes to take training steps faster. Table \ref{vertab2} shows the training speed of the FNet and our proposed models for [Model]-S and [Model]-M model specifications during the pretraining process. The results suggest that our proposed methods with integrated pooling layers work faster than the FNet and the Fast-FNet variant with a dense layer, which is expected since the dense layer adds additional computations both for feed-forward and back-propagation stages. One should also note that, although FNet is faster than the proposed Fast-FNet with the Max Pooling when using $4$ encoder blocks, i.e. [Model]-S specifications, its training speed decreases more than others and becomes slower than that of the Max Pooling variant while increasing the number of encoder blocks. Considering that the [Model]-Base specification of BERT model uses $12$ encoder blocks, which is the standard number, the difference in training speed may become more obvious if we use $12$ encoder blocks. Considering the pretraining process lasts for $1$ Million steps, the difference between reported step per second values becomes more valuable.  

\begin{table}[htb]
\centering 
\begin{tabular}{lcc}
\toprule   

\multirow{2}{*}{\textbf{Model}}  & \multicolumn{2}{c}{\textbf{Model Spec.}}\\
&-S&-M\\
\hline
FNet &  2.19 &  2.06\\ 
Fast-FNet MaxP & 2.17 &  2.14\\ 
Fast-FNet MeanP & \textbf{2.23} &  \textbf{2.20}\\
Fast-FNet Dense& 2.10 &  2.02\\
\bottomrule
\end{tabular}

\vspace{0.5em}
\caption{Training steps per second of the proposed models in comparison with the FNet. } 
\label{vertab2}
\end{table}

Upon observing Tables \ref{paramsize} and \ref{vertab2}, the gain in efficiency is obvious. Our proposed architectures occupy less memory since they have less parameters than the FNet. Moreover, our proposed models work faster and allow us to save considerable amounts of time during training. Although these aforementioned advantages are quite valuable in such an era where the deep learning models require enormous computational burdens, our proposed models have to also maintain comparable performances with computationally more expensive models. In other words, the gain in training efficiency should not sacrifice model performances notably.

\begin{table*}[!h]
\centering 
\resizebox{1.8\columnwidth}{!} {%
\begin{tabular}{l@{\extracolsep{4pt}}cccccccc}
\toprule   

\textbf{Model}  & \textbf{MNLI}  & \textbf{QQP} & \textbf{QNLI} & \textbf{SST-2} & \textbf{CoLA} & \textbf{STS-B} & \textbf{MRPC} & \textbf{RTE}\\
\midrule
FNet ``S''  & 61.7/62.3 & \textbf{80.9}  & 74.5 & 78.4 & \textbf{69.1} & \textbf{70.7} & \textbf{76.5} & 54.5 \\ 
Fast-FNet MaxP-S & 61.3/62.3 & 76.9  & 72.7 & 79.4 & \textbf{69.1} & 64.4 & 74.5 & 55.2 \\
Fast-FNet MeanP-S & 62.0/\textbf{63.6} & 80.0  & 71.6 & \textbf{79.5} & \textbf{69.1} & 65.8 & 73.7 & \textbf{57.0} \\
Fast-FNet Dense-S & \textbf{62.3}/63.5 & 79.5  & \textbf{74.6} & 79.2 & 69.0 & 67.2 & 74.3 & 56.3 \\ \hline
FNet-M  & \textbf{67.8/69.5}  & 80.3  & 77.5 & 82.3 & 69.1 & 76.0 & \textbf{76.5} & 53.4 \\ 
Fast-FNet MaxP-M & 65.5/67.7 & \textbf{82.2}  & 79.7 & 81.7 & 69.1 & 74.6 & 74.8 & 54.9 \\
Fast-FNet MeanP-M & 66.1/67.8 & 81.5  & \textbf{80.0} & \textbf{82.7} & 69.1 & \textbf{78.6} & 74.4 & 58.1 \\
Fast-FNet Dense-M & 65.5/67.5 & 80.5  & 78.0 & 82.6 & 69.1 & 73.0 & 76.2 & \textbf{58.5} \\
\bottomrule
\end{tabular}
}
\vspace{0.5em}
\caption{GLUE validation results for the [Model]-S and [Model]-M model specifications. The mean of accuracy and F1 scores for QQP and MRPC tasks, the Spearman correlation for STS-B, and accuracy scores for other tasks are reported. The scores for both matched and mismatched versions of the MNLI dataset are given.}
\label{vertab1}
\end{table*}

As we mentioned before, \cite{lee2021fnet} has already shown that the performance of FNet on GLUE tasks is still comparable with BERT, although it works quite faster. Having shown that our proposed models are even faster than FNet, we also conducted experiments on GLUE tasks to demonstrate that our models do not sacrifice performance on this benchmark. The results for GLUE tasks are provided in Table \ref{vertab1}, where we reported our results for both [Model]-S and [Model]-M model specifications. The models are trained on the C4 dataset for $1$ Million training steps and fine-tuned for GLUE tasks. The results suggest that, our models can still perform on par with FNet and even outperform FNet on some tasks. Specifically, on [Model]-S specification, only relatively significant difference is the performance of models on ``STS-B'' task. However, FNet model outperformed the best of Fast-FNet model, which is the ``Dense'' variation, only by $3.5\%$. On the other hand, on [Model]-M specification, Fast-FNet variations mostly outperformed FNet and even when they are outperformed, their performance remained competitive regarding FNet's results.

\subsubsection{\textbf{LRA}}
For the tasks in LRA benchmark, we performed two sets of experiments. In the first one, where we compared the Fourier-based Transformer models with other attention-based Transformer models with different attention types, we used the same model configurations as provided in \cite{chen21skyformer}. However, to fit everything into one GPU and complete all experiments with the same computational resources, we decreased batch sizes for some tasks. More specifically, we decreased batch size value from $16$ to $8$ for the ``Retrieval'' task, from $32$ to $8$ for the ``Text'' task and from $256$ to $128$ for the ``Image'' task. 

Then, in the second part, where we specifically compared the efficiency of FNet and our proposed models, we increased the hidden dimension of Transformer and conduct our experiments with changing number of layers. 

Note that, all the results are reproduced for every attention-based Transformer model using the specified computational resources in order to make a clear comparison of performance and efficiency.


The accuracy, training speed, and memory usage results for the first part of the experiments are given in Tables \ref{lraaccresults}, \ref{lratraintime}, and \ref{lramemory}, respectively. In Table \ref{lraaccresults}, the accuracy results for each task are given. Here, the attention-based Transformer models with more complicated and computationally expensive attention mechanisms are mostly gave the best results. However, both FNet's and our proposed models' performances are comparable with other models. Although the FNet and our proposed models could not give the best performance, at least one of them could outperform or at least stay close to most of the attention-based Transformer models.

\begin{table}[htb]
\centering
\resizebox{\columnwidth}{!}{
\begin{tabular}{lccccc}
\hline
\multicolumn{1}{c}{\multirow{2}{*}{\textbf{Model}}}                                                                                                               & \multicolumn{5}{c}{\textbf{Task}}                                                                                                                                  \\ \cline{2-6} 
\multicolumn{1}{c}{}                                                                                                                                              & \textbf{ListOps}               & \textbf{Pathfinder}            & \textbf{Text}                  & \textbf{Retrieval}             & \textbf{Image}                 \\ \hline
\textbf{FNet}& 37.5 (-1.7\%)  & 51 (-20.8\%)  & 61.7 (\textbf{-3.5\%})  & 74.5 (-6.4\%)   & 26.1 (-17.3\%)  \\ \hline

\textbf{\begin{tabular}[c]{@{}l@{}}Fast-FNet\\ MaxP\end{tabular}} &   37.7 (-1.5\%)             & 53.4 (\textbf{-18.4\%})                  & 60.6 (-4.6\%)          & 76.0 (-4.9\%)                 & 28.6 (-14.8\%)           \\ \hline

\textbf{\begin{tabular}[c]{@{}l@{}}Fast-FNet\\ MeanP\end{tabular}} & 38.5 (\textbf{-0.7\%})             & 50.8 (-19.0\%)                  & 60.7 (-4.5\%)          & 76.0 (-4.9\%)                 & 29.0 (-14.4\%)           \\ \hline

\textbf{\begin{tabular}[c]{@{}l@{}}Fast-FNet\\ Dense\end{tabular}}       & 37.8 (-1.4\%)             & 51.9 (-19.9\%)                & 61.5 (-3.7\%)          & 76.1 (\textbf{-4.8\%})               & 33.9 (\textbf{-9.5\%})           \\ \hline\hline

\textbf{Transformer}                                                    & 38.9             & \textbf{71.8}                & 61.8          & 78.4               & 40.5           \\ 
\textbf{Skyformer}                                                      & \textbf{39.2}             & 71                  & 60.8          & \textbf{80.9}               & 33.8           \\ 
\textbf{Linformer}                                                      & 37.3             & 56.4                & 60            & 76.1               & 38             \\ 
\textbf{Informer}                                                      & 24.5             & 56                & 63.7            & 72.2               & 38.1             \\ 
\textbf{Performer}                                                      & 38.1             & 68.8                & 63.7            & 78.5               & 38.2             \\ 
\textbf{Reformer}                                                      & 23.9             & 65.4                & 64.6            & 78.1               & \textbf{43.4}             \\ 
\textbf{Bigbird}                                                      & 38.4             & 70                & 63.3            & 78.5               & 39.6             \\ 
\textbf{Nystrom}                                                        & 38.3             & 70.2                & \textbf{65.2}          & 80                 & 40             \\ 

\hline

\end{tabular}}
\vspace{0.2em}
\caption{LRA benchmark test accuracies (\%). For the Fourier-based models, the loss in accuracy compared to the best performing attention-based Transformer model is given.}
\label{lraaccresults}
\end{table}

In Table \ref{lratraintime}, the number of steps that each model can take in a second in given for each task. Looking at the averages, we can see that the best of Fourier-based Transformer model, the Fast-FNet with max pooling, can outperform the best of the attention-based Transformer model, ``Linformer'', by $91\%$ and the difference climbs up to about 7-fold when considering the attention-based Transformer models with worse efficiency. These results show that the difference between attention-based Transformer models with expensive attention mechanisms and Fourier-based Transformer models in efficiency is immensely high. Moreover, the Fourier-based models do not lose significant model performance as shown in the previous experimental results.

\begin{table}[htb]
\centering
\resizebox{\columnwidth}{!}{
\begin{tabular}{lccccc|c}
\hline
\multicolumn{1}{c}{\multirow{2}{*}{\textbf{Model}}}                                                                                                               & \multicolumn{6}{c}{\textbf{Task}}                                                                                                                                  \\ \cline{2-7} 
\multicolumn{1}{c}{}                                                                                                                                              & \textbf{ListOps}               & \textbf{Pathfinder}            & \textbf{Text}                  & \textbf{Retrieval}             & \textbf{Image}    & \textbf{Average}                   \\ \hline 

\textbf{FNet}                                                                        & 20.55             & 10.67                & 33.33          & 19.47               & 10.31      & 18.87     \\ \hline

\textbf{\begin{tabular}[c]{@{}l@{}}Fast-FNet\\ MaxP\end{tabular}}              & \textbf{21.83}             & \textbf{10.94}                & \textbf{35}          & 19.23               & \textbf{10.82}      & \textbf{19.56}     \\ \hline

\textbf{\begin{tabular}[c]{@{}l@{}}Fast-FNet\\ MeanP\end{tabular}}             & 21.32             & 10.86                & 33.74          & \textbf{20.29}               & 10.7       & 19.38    \\ \hline

\textbf{\begin{tabular}[c]{@{}l@{}}Fast-FNet\\ Dense\end{tabular}}                     & 21.53             & 10.46                & 34.11          & 16.86               & 10.48      & 18.69     \\ \hline\hline

\textbf{Transformer}                                                                 & 2.89            & 2.3               & 2.97         & 1.55              & 2.39     & 2.42     \\ 
\textbf{Skyformer}                                                                   & 1.79            & 1.85               & 6.99          & 3.74              & 1.79    & 3.23      \\ 
\textbf{Linformer}                                                                   & 11.31             & 5.85                & 17.91          & 10.41               & 5.7 &10.24          \\ 
\textbf{Informer}                                                                    & 5.21             & 2.4                & 7.46          & 4.19               & 2.28   &4.31        \\ 
\textbf{Performer}                                                                   & 8.39             & 3.86                & 13.93            & 7.77               & 4.17      &7.62       \\ 
\textbf{Reformer}                                                                    & 7.64             & 3.76                & 15.5            & 7               & 3.69    &7.52         \\ 
\textbf{Bigbird}                                                                     & 3.62             & 1.91                & 6.17            & 3.28               & 1.87     &3.37        \\ 
\textbf{Nystrom}                                                                     & 9.21             & 4.25               & 16.29          & 8.87               & 4.14    &8.55      \\ \hline

\end{tabular}}
\vspace{0.2em}
\caption{LRA number of steps taken in a second, averaged over 50K training steps.}
\label{lratraintime}
\end{table}

Besides the increase in the training speed of Fourier-based Transformer models, the proposed Fast-FNet Transformer models require much less computational memory allocation, which is also another indicator of efficiency. In Table \ref{lramemory}, the peak usages of GPU memory are reported. Here, the difference between Fourier-based Transformer models and attention-based Transformer models are even more noticeable. The most memory-efficient attention-based Transformer model, ``Linformer'', allocates over $3$ times more memory in average than the most memory-efficient Fourier-based Transformer model, the Fast-FNet with mean pooling, and the worst memory-efficient model, ``Transformer'', allocates almost $14$ times more memory. The striking difference between the Fourier-based Transformer models and the attention-based full Transformer models are caused by the removal of weight matrices that are present in attention mechanisms. Getting rid of the learning procedures for those computationally expensive matrices and the matrix operations, the Fourier-based Transformer models can fasten the training and require less GPU allocation.

\begin{table}[htb]
\centering
\resizebox{\columnwidth}{!}{
\begin{tabular}{lccccc|c}
\hline
\multicolumn{1}{c}{\multirow{2}{*}{\textbf{Model}}}                                                                                                               & \multicolumn{6}{c}{\textbf{Task}}                                                                                                                                  \\ \cline{2-7} 
\multicolumn{1}{c}{}                                                                                                                                              & \textbf{ListOps}               & \textbf{Pathfinder}            & \textbf{Text}                  & \textbf{Retrieval}             & \textbf{Image}    & \textbf{Average}                   \\ \hline

\textbf{FNet}                                                           & 342              & 682                 & 175           & 331                & 681     &    442   \\ \hline

\textbf{\begin{tabular}[c]{@{}l@{}}Fast-FNet\\ MaxP\end{tabular}} & 328              & 653                 & 167           & 316                & 653      &   423   \\ \hline

\textbf{\begin{tabular}[c]{@{}l@{}}Fast-FNet\\ MeanP\end{tabular}} & \textbf{312}              & \textbf{621}                 & \textbf{159}           & \textbf{300}                & \textbf{621}       &   \textbf{402}  \\ 
\hline
\textbf{\begin{tabular}[c]{@{}l@{}}Fast-FNet\\ Dense\end{tabular}}       & 328              & 653                 & 167           & 316                & 653     &    423   \\ \hline\hline

\textbf{Transformer}                                                    & 5498             & 5879                & 5313          & 5518               & 5873     &   5616   \\ 
\textbf{Skyformer}                                                      & 1790             & 4229                & 817           & 1616               & 4229        &  2536 \\ 
\textbf{Linformer}                                                      & 1010             & 2019                & 514           & 976                & 2019       &  1308  \\ 
\textbf{Informer}                                                      & 4969             & 4869                & 2627           & 2961                & 4869      &   4059  \\
\textbf{Performer}                                                      & 1117             & 2235                & 562            & 1105               & 2234      & 1451      \\ 
\textbf{Reformer}                                                      & 1647             & 3290                & 827            & 1531               & 3290      & 2117      \\ 
\textbf{Bigbird}                                                      & 2779             & 5094                & 1453            & 2555               & 5093     &3395        \\ 
\textbf{Nystrom}                                                        & 1399             & 3435                & 623           & 1228               & 3434    &  2024     \\ \hline

\end{tabular}}
\vspace{0.2em}
\caption{LRA peak memory usage (MBs).}
\label{lramemory}
\end{table}

Although we have shown that the Fourier-based Transformer models are more efficient than attention-based Transformer models, the results that are given in Tables \ref{lratraintime} and \ref{lramemory} do not indicate a clear difference between the FNet and our proposed models. Despite the slight efficiency improvement of our proposed models compared to FNet, since the provided model configurations suggest us to use only $2$ encoder blocks, the difference seems to be insignificant. 

\begin{figure}[h]
\centering
\begin{subfigure}{.5\linewidth}
  \centering
  \includegraphics[scale=0.3]{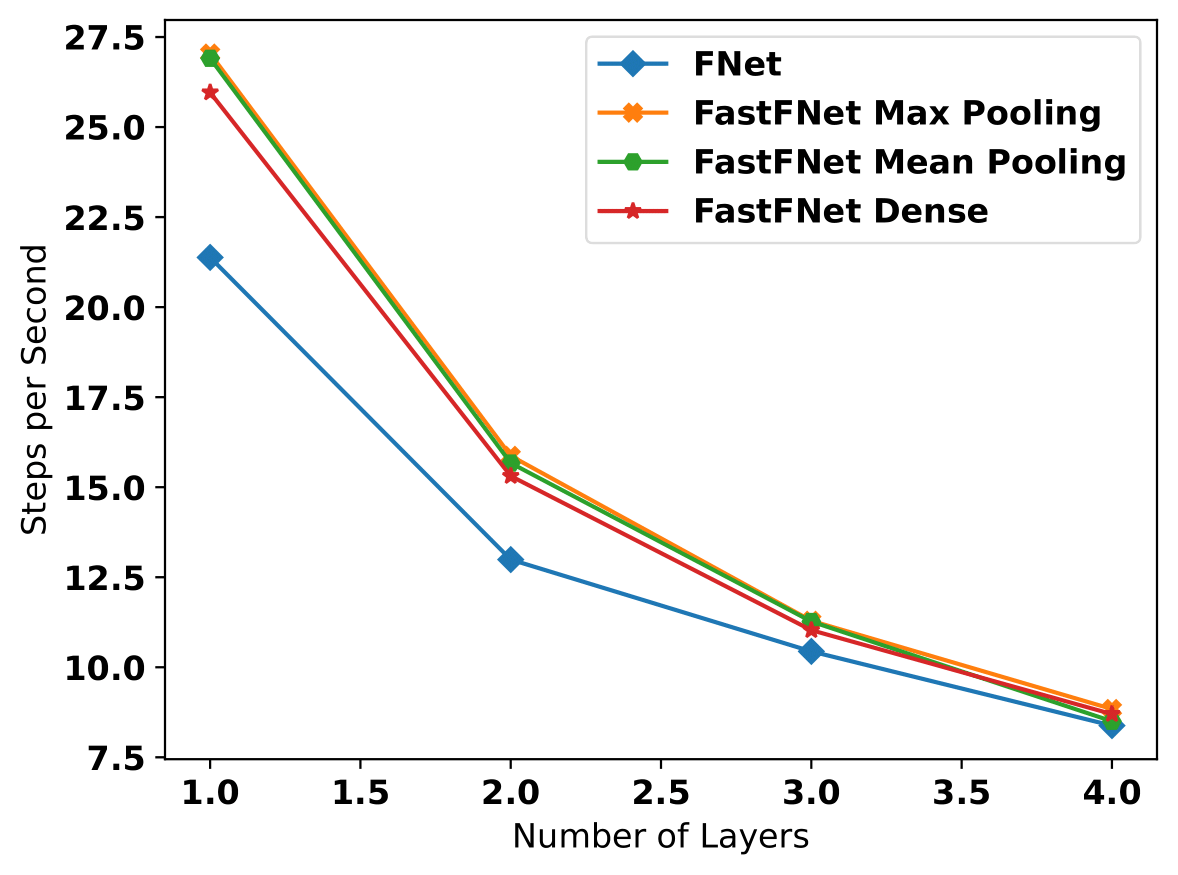}
  \caption{ListOps}
\end{subfigure}%
\begin{subfigure}{.5\linewidth}
  \centering
  \includegraphics[scale=0.3]{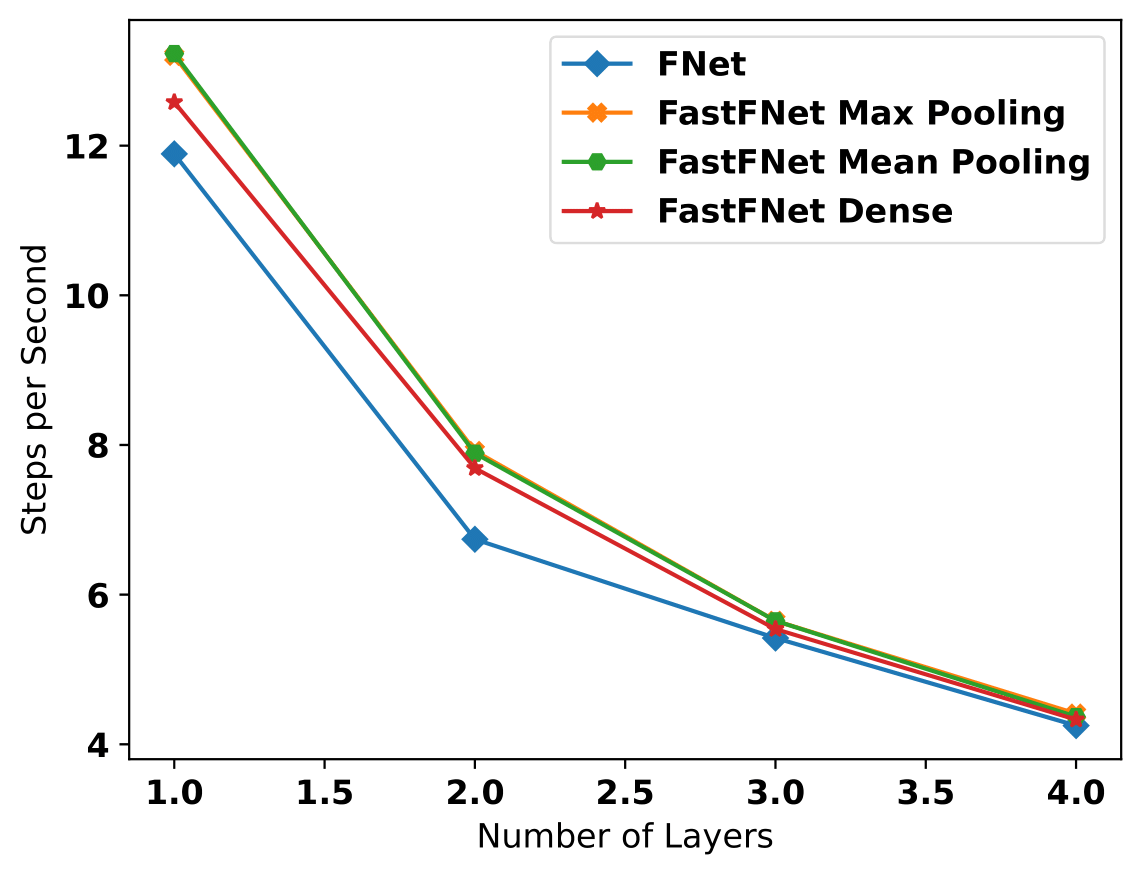}
  \caption{Pathfinder}
\end{subfigure}
\begin{subfigure}{.5\linewidth}
  \centering
  \includegraphics[scale=0.3]{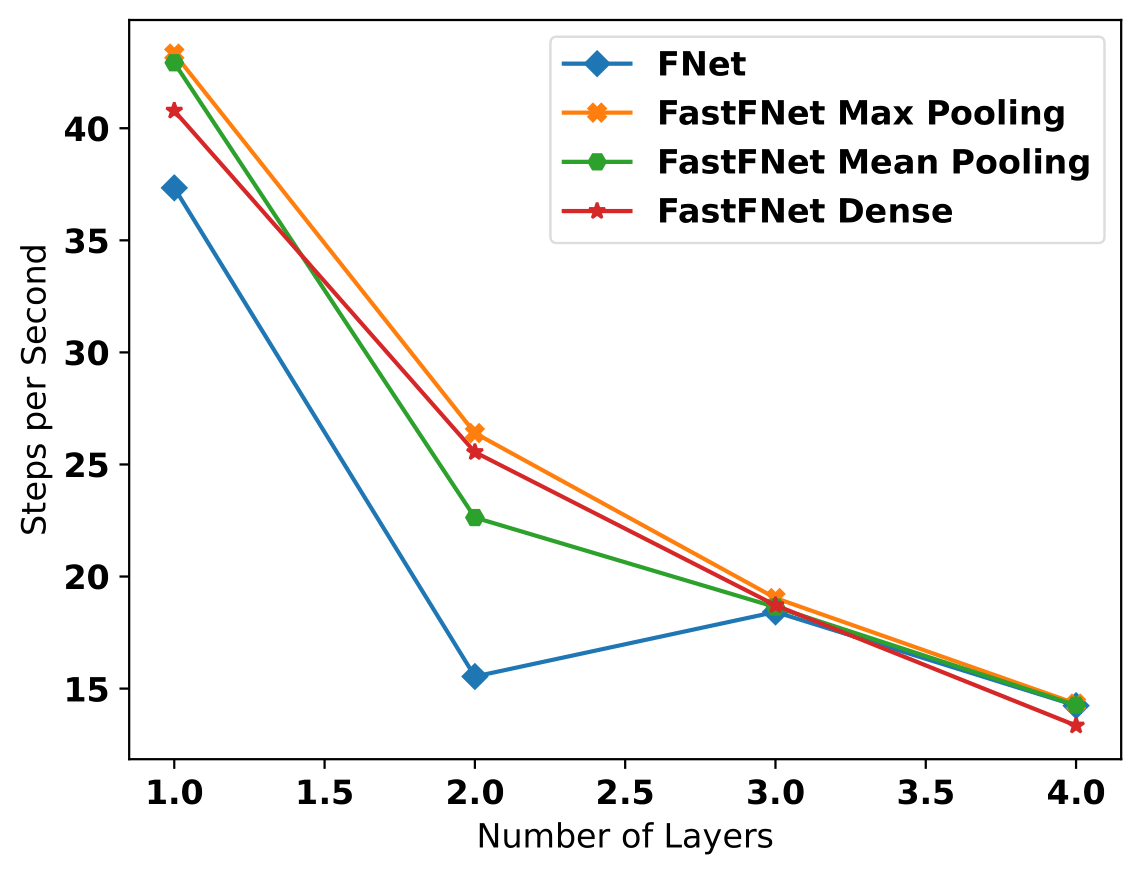}
  \caption{Text}
\end{subfigure}%
\begin{subfigure}{.5\linewidth}
  \centering
  \includegraphics[scale=0.3]{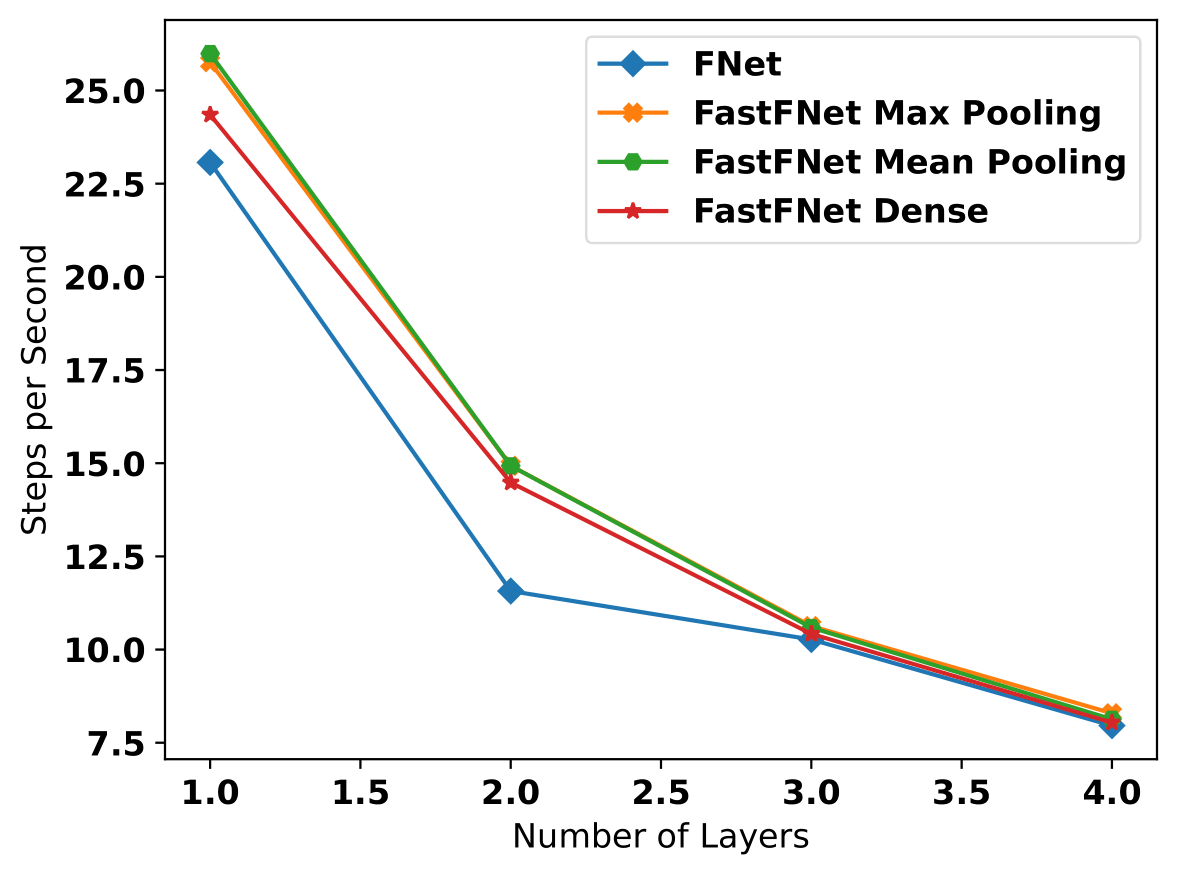}
  \caption{Retrieval}
\end{subfigure}
\begin{subfigure}{.5\linewidth}
  \centering
  \includegraphics[scale=0.3]{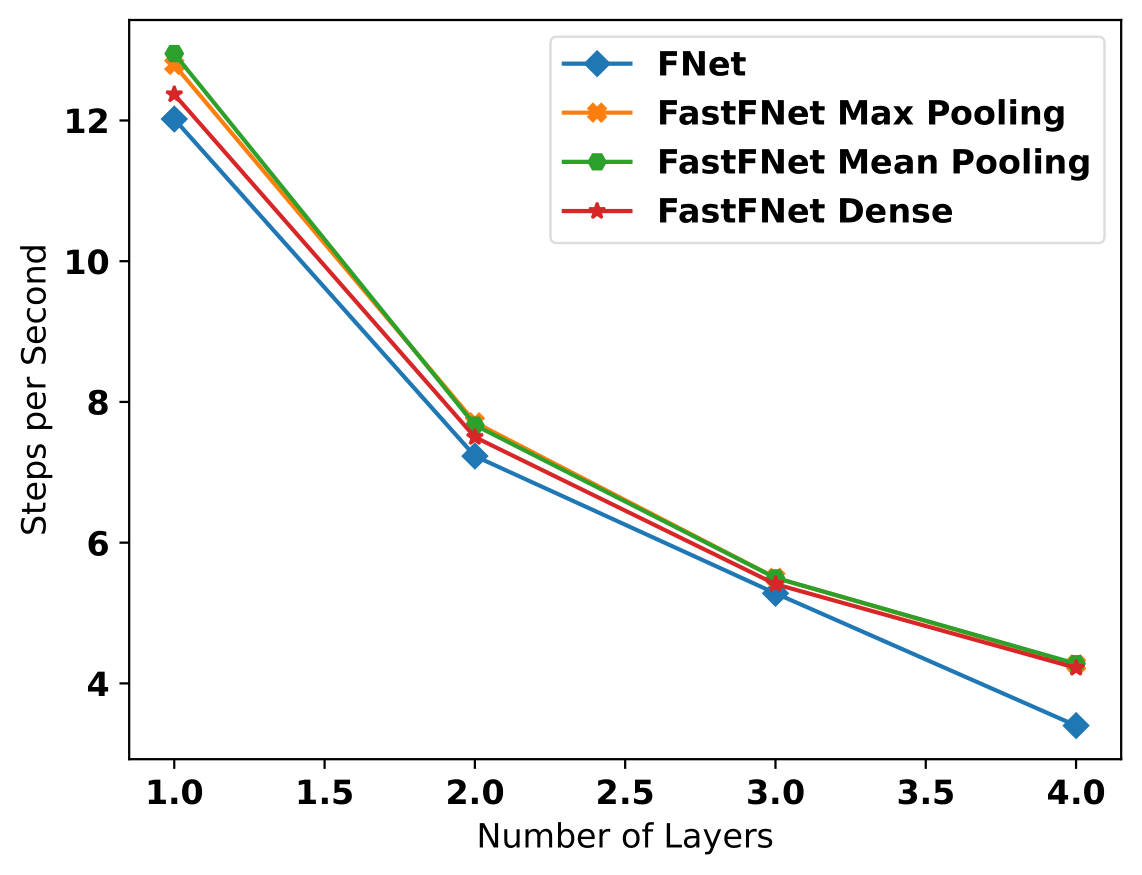}
  \caption{Image}
\end{subfigure}%
\begin{subfigure}{.5\linewidth}
  \centering
  \includegraphics[scale=0.3]{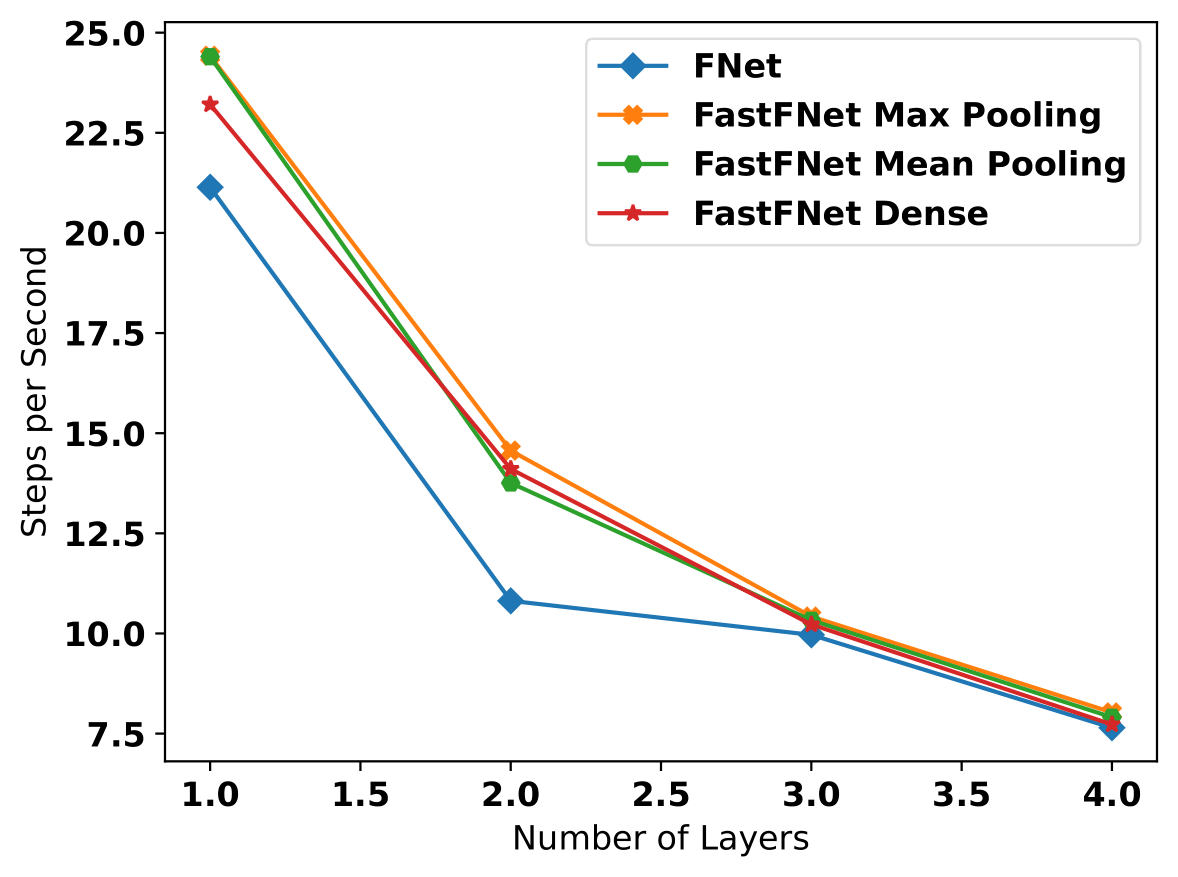}
  \caption{Average}
\end{subfigure}
\caption{The number of steps taken in a second for each LRA task.}
\label{lraspefig}
\end{figure}

\begin{figure}[h]
\centering
\begin{subfigure}{.5\linewidth}
  \centering
  \includegraphics[scale=0.3]{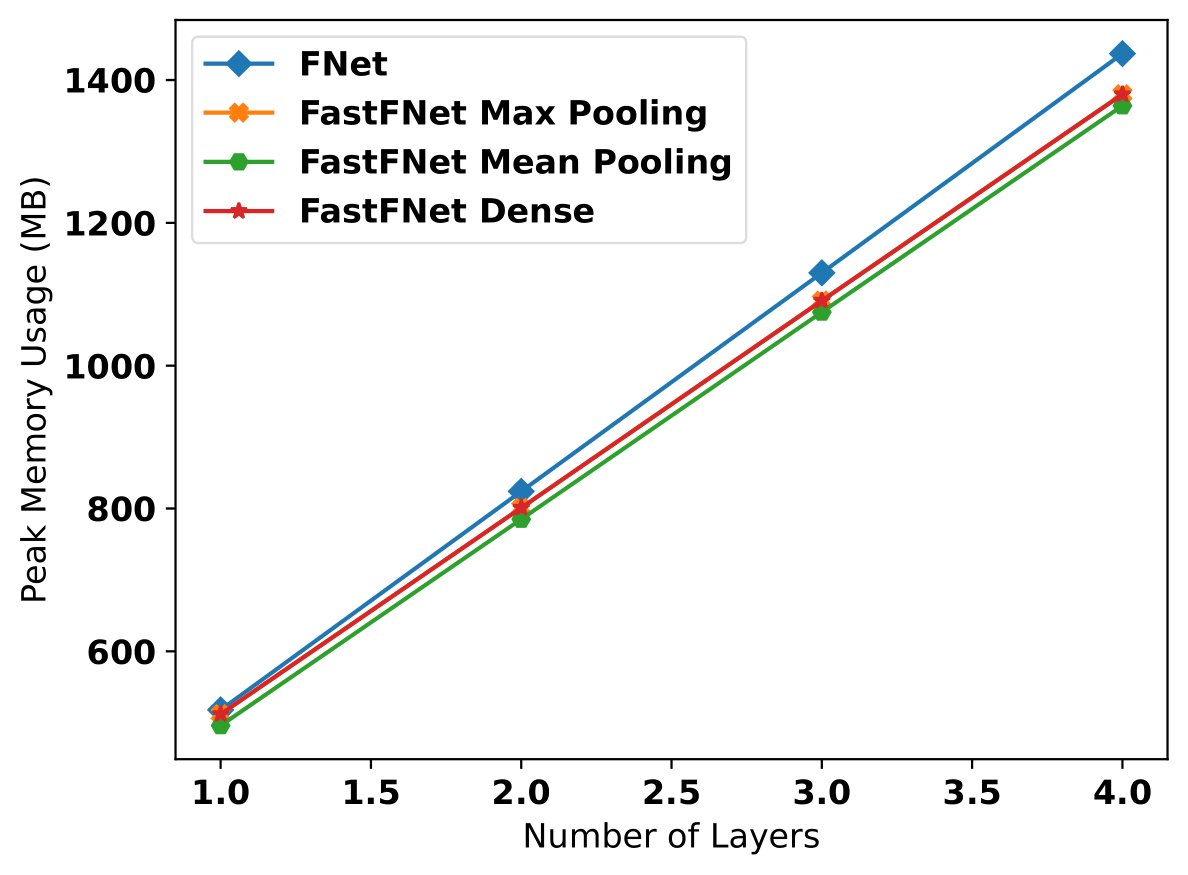}
  \caption{ListOps}
\end{subfigure}%
\begin{subfigure}{.5\linewidth}
  \centering
  \includegraphics[scale=0.3]{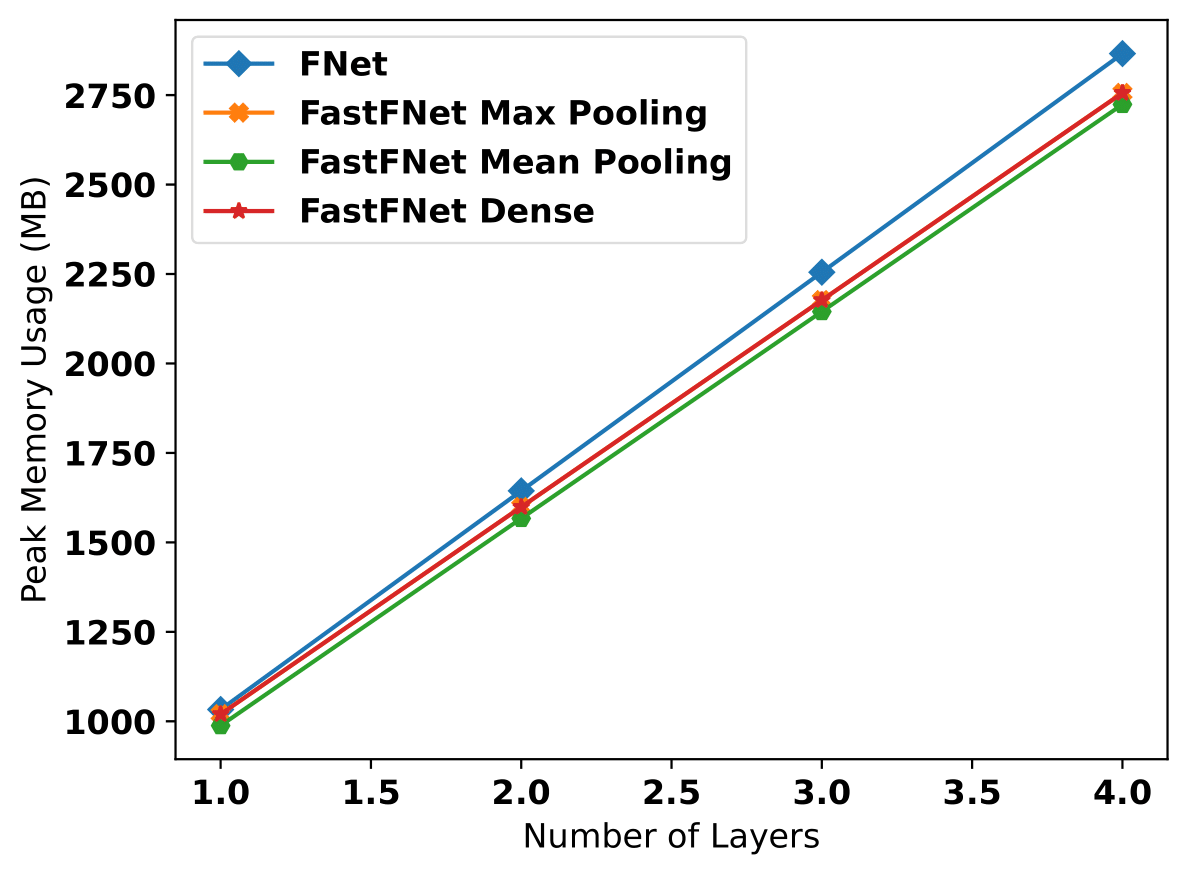}
  \caption{Pathfinder}
\end{subfigure}
\begin{subfigure}{.5\linewidth}
  \centering
  \includegraphics[scale=0.3]{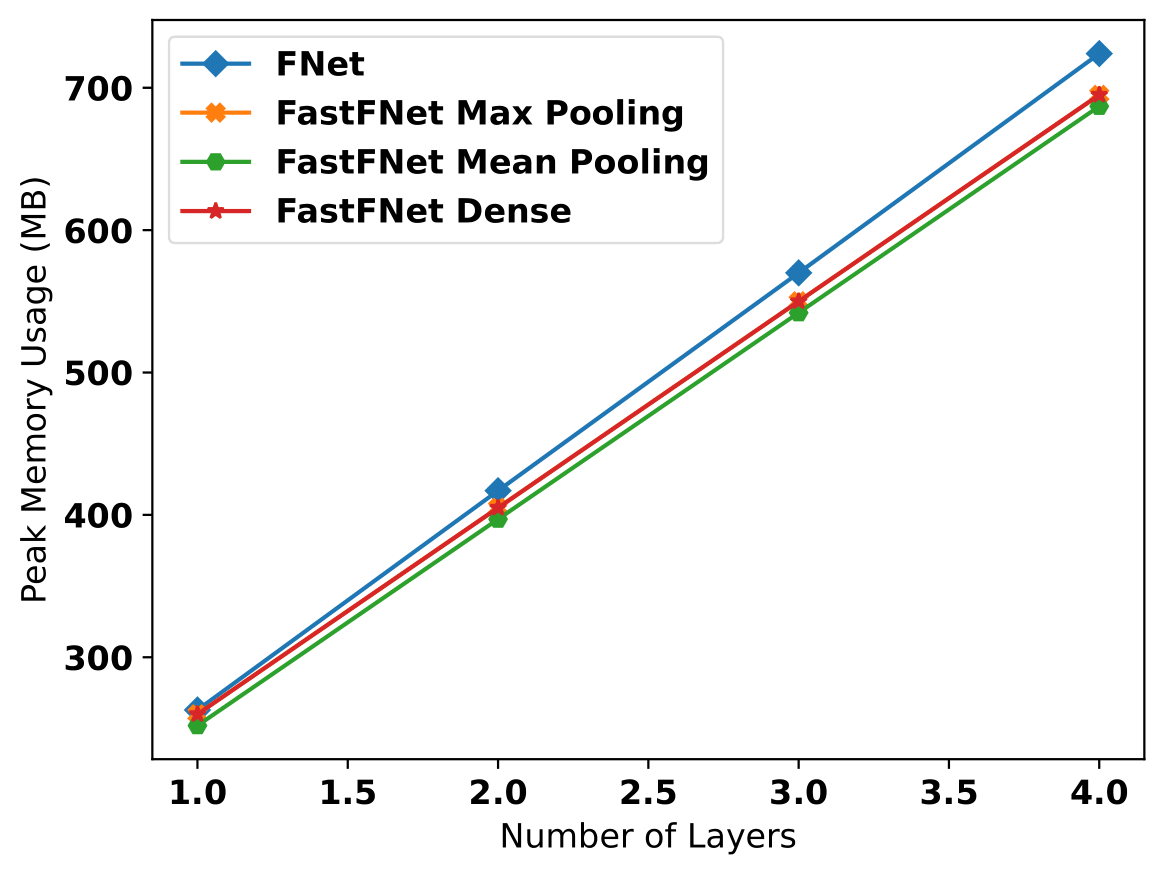}
  \caption{Text}
\end{subfigure}%
\begin{subfigure}{.5\linewidth}
  \centering
  \includegraphics[scale=0.3]{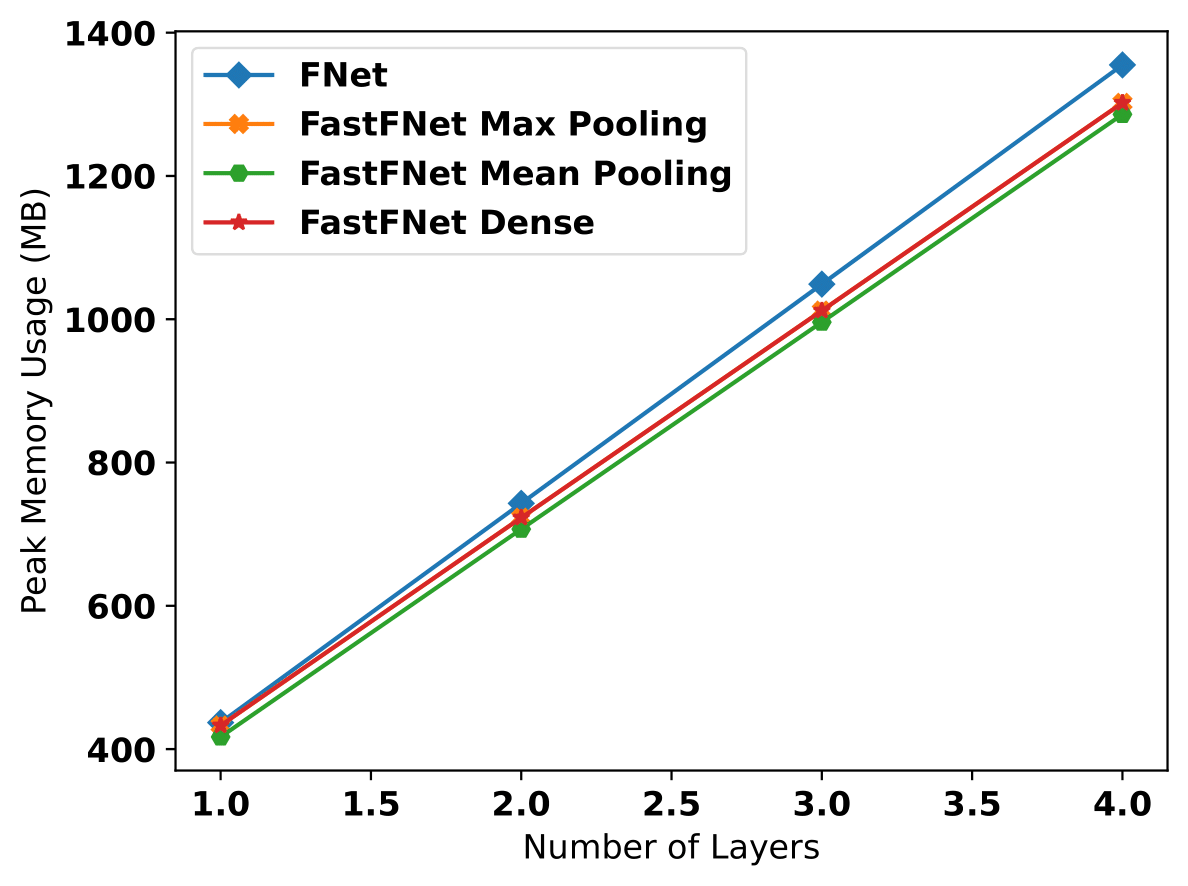}
  \caption{Retrieval}
\end{subfigure}
\begin{subfigure}{.5\linewidth}
  \centering
  \includegraphics[scale=0.3]{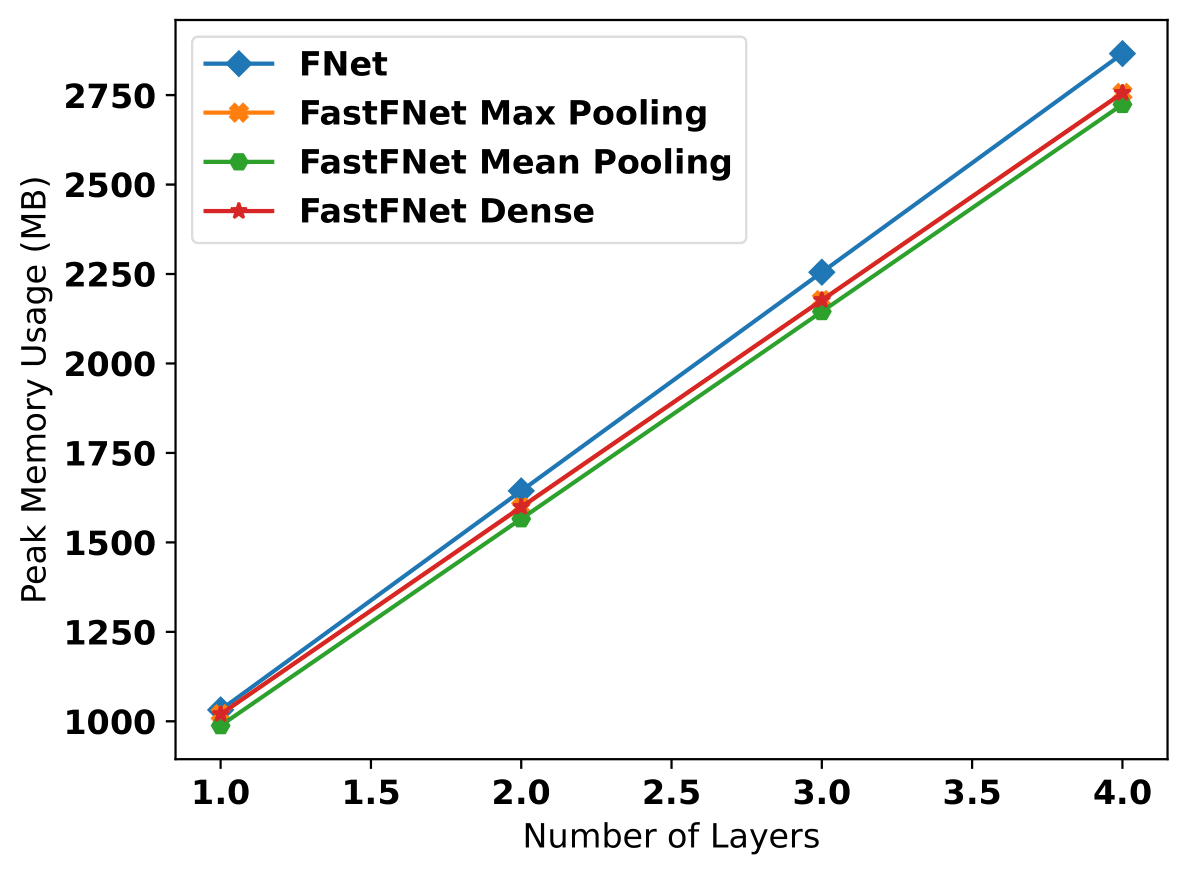}
  \caption{Image}
\end{subfigure}%
\begin{subfigure}{.5\linewidth}
  \centering
  \includegraphics[scale=0.3]{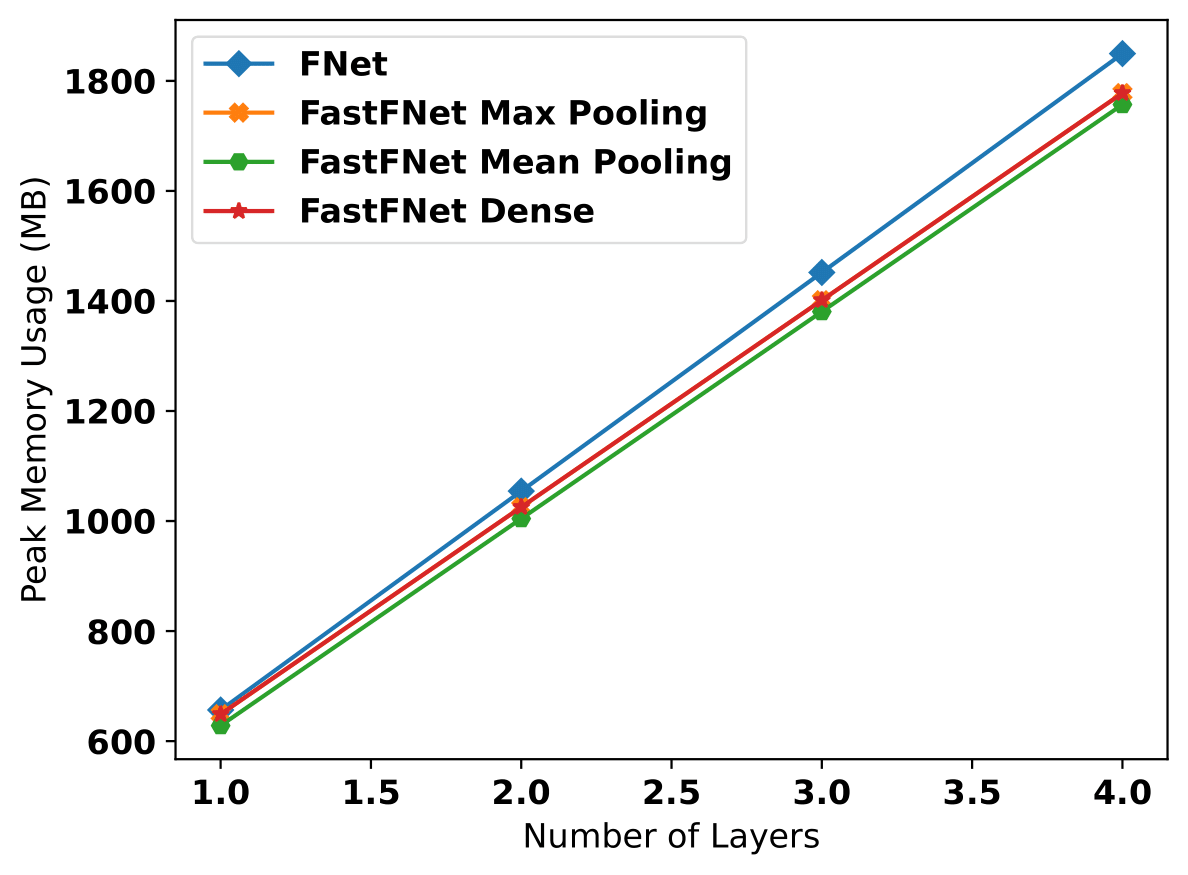}
  \caption{Average}
\end{subfigure}
\caption{The peak value of memory allocation for each LRA task.}
\label{lramemfig}
\end{figure}

As a further study to observe the effect of the number of encoder blocks used, we also experimented our proposed models in comparison with the FNet by using larger hidden dimension and increasing number of encoder blocks. More specifically, we increased the hidden dimension from $64$ to $512$ and experimented with $1$, $2$, $3$, and $4$ encoder blocks. The results for training speed and memory usage are given in Figs. \ref{lraspefig} and \ref{lramemfig}, respectively. For the both figures, the first five sub-figures show the results of LRA tasks, i.e ListOps, Pathfinder, Text, Retrieval, and Image, and the last sub-figure is the average of those tasks. The results show that our proposed models are working faster than the FNet while using less GPU memory. The difference in training speed is more obvious with less number of layers and every model converges to similar values in terms of steps taken in a second. The difference in memory usage, however, increases constantly while the number of layers increases and the gap between the Fast-FNet and FNet models becomes more evident.

\section{Conclusion}\label{sec:conclusion}

In this work, we proposed an efficient Transformer architecture that replaces the burdensome attention layer with a modified FT layer. Leveraging FT properties from signal processing, we removed the redundant information and reduced the hidden dimensions in the Transformer model. We replaced the attention layer with a special Fourier layer where only the half of real parts of the Fourier-transformed input data is taken. To avoid dimensionality mismatches, we also reduced the dimension of the residual connections. To this end, we used max pooling, mean pooling and a standard dense layer for the reduction of residual connections. 

We tested our model mainly in comparison with the FNet architecture and proved that the reduction in dimensionality brought gain in both time and GPU memory usage. Although the dimensionality inside the Transformer is reduced and, thus, the training became faster and less expensive in terms of computational power, we still maintain the same, and sometimes better, performance in numerous benchmark tasks. 

Considering that the deep learning models are computationally getting more and more expensive recently and the majority of researchers may not possess powerful machines to train such models, it is crucial to have alternatives that ease the burden of heavy models. Besides, considering the environmental effects of computationally expensive deep learning models, where Transformer architecture is one of the most expensive one among all, we find useful to prefer models that require less memory allocation and last shorter to train in order to reduce the carbon footprints.


\ifCLASSOPTIONcaptionsoff
  \newpage
\fi

\balance
\bibliographystyle{IEEEtran}
\bibliography{bare_jrnl}

\end{document}